\documentclass[preprint,times,11pt]{elsarticle}

\usepackage{lineno,hyperref}
\usepackage{amsfonts}
\usepackage{amsmath}
\usepackage{mathtools}
\usepackage{multirow}
\usepackage{comment}
\usepackage{array}
\usepackage{diagbox}
\usepackage{color}
\usepackage[rightcaption]{sidecap}

\modulolinenumbers[5]
\newcommand{\etal}{\textit{et al.}}
\journal{Journal of \LaTeX Templates}

\newif\ifdraft
\drafttrue

\ifdraft
     \newcommand{\liu}[1]{\textcolor{blue}{{[liu: #1]}}}
\else
    \newcommand{\liu}[1]{}
\fi

\ifdraft
     \newcommand{\wm}[1]{\textcolor{magenta}{{[wwm: #1]}}}
\else
    \newcommand{\wm}[1]{}
\fi

\begin{document}

\begin{frontmatter}

\title{Enhancing Local Feature Learning for 3D Point Cloud Processing using Unary-Pairwise Attention}

\author[titech_es,aist_airc]{Haoyi Xiu}
\ead{xiu.h.aa@m.titech.ac.jp; hiroki-shuu@aist.go.jp}

\author[aist_airc]{Xin Liu \corref{cor}}
\ead{xin.liu@aist.go.jp}

\author[dlut]{Weimin Wang \corref{cor}}
\ead{wangweimin@dlut.edu.cn}

\author[aist_airc]{Kyoung-Sook Kim}
\ead{ks.kim@aist.go.jp}

\author[titech_es]{Takayuki Shinohara}
\ead{shinohara.t.af@m.titech.ac.jp}

\author[titech_cs]{Qiong Chang}
\ead{q.chang@c.titech.ac.jp}

\author[titech_es,aist_airc]{Masashi Matsuoka}
\ead{matsuoka.m.ab@m.titech.ac.jp}

\address[titech_es]{Department of Architecture and Building Engineering, Tokyo Institute of Technology, Tokyo, Japan}
         
\address[aist_airc]{Artificial Intelligence Research Center, AIST, Tokyo, Japan}

\address[dlut]{DUT-RU International School of Information Science and Engineering, Dalian University of Technology, Dalian, China}


\address[titech_cs]{Department of Computer Science, Tokyo Institute of Technology, Tokyo, Japan}

\cortext[cor]{Corresponding authors.}

\begin{abstract}
We present a simple but effective attention named the unary-pairwise attention (UPA) for modeling the relationship between 3D point clouds. Our idea is motivated by the analysis that the standard self-attention (SA) that operates globally tends to produce almost the same attention maps for different query positions, revealing difficulties for learning query-independent and query-dependent information jointly. Therefore, we reformulate the SA and propose query-independent (Unary) and query-dependent (Pairwise) components to facilitate the learning of both terms. In contrast to the SA, the UPA ensures query dependence via operating locally. Extensive experiments show that the UPA outperforms the SA consistently on various point cloud understanding tasks including shape classification, part segmentation, and scene segmentation. Moreover, simply equipping the popular PointNet++ method with the UPA even outperforms or is on par with the state-of-the-art attention-based approaches. In addition, the UPA systematically boosts the performance of both standard and modern networks when it is integrated into them as a compositional module.

\end{abstract}






\end{frontmatter}


\section{Introduction}
\label{sec:intro}
3D data has become increasingly available thanks to the advent of modern 3D sensors. The 3D point cloud is one of the simplest shape representations which is typically represented as spatially scattered 3D points. Recently, automatic understanding of 3D point clouds using deep learning~\cite{lecun2015deep} has attracted much interest in various applications such as autonomous driving~\cite{cui2020deep, qi2018frustum} and remote sensing~\cite{zhu2017deep,shinohara2020fwnet}.

The irregular nature of 3D point clouds brings challenges on the deep learning--based point cloud analysis because popular methods like convolutional neural network (CNN) only work on the regularly structured data (e.g., 2D and 3D grids). Therefore, 3D point clouds are often projected to regular formats such as voxels~\cite{maturana2015voxnet, zhou2018voxelnet} and images~\cite{kanezaki2018rotationnet, su2015multi} to enable regular convolutions. Recently, PointNet~\cite{qi2017pointnet} has triggered the development of methods that directly operate on point clouds~\cite{qi2017pointnet++, wang2019dynamic, thomas2019kpconv}. The key to their success is rooted in the use of shared multi-layer perceptrons (MLPs) and symmetric functions (e.g., max-pooling and avg-pooling). Both types of operations ensure permutation invariance, making them a perfect fit for point cloud processing.

On the other hand, the success of the self-attention (\textbf{SA})~\cite{vaswani2017attention, devlin2018bert} in natural language processing has triggered various applications of the SA to 2D vision problems (e.g., image recognition~\cite{bello2019attention}, generation~\cite{zhang2019self}, and object detection~\cite{wang2018non}). The SA updates the query features by aggregating features from other positions (keys) based on pairwise relationships. The SA is permutation invariant; therefore, it is directly applicable to 3D point clouds. Recent research shows that the SA can indeed benefit the point cloud processing~\cite{lee2019set,yang2019modeling,guo2020pct}.
\begin{figure}[b]
    \begin{center}
        \includegraphics[width=0.9\linewidth]{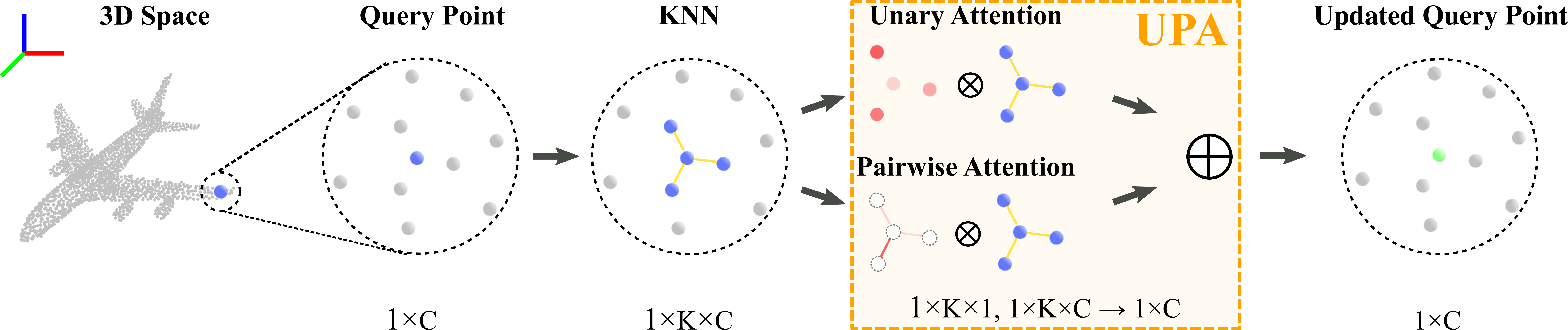}
    \end{center}
    \caption{\textbf{Overview of UPA.} Given a query point (the blue point), UPA updates the feature by combining the outputs from unary and pairwise attentions. $\bigotimes$ represents matrix multiplication and $\bigoplus$ denotes element-wise addition.}
    \label{fig: overview}
\end{figure}
However, it is found in the 2D vision domain that the SA often produces almost the same attention maps for very different query positions~\cite{cao2019gcnet,yin2020disentangled}. Such a finding is crucial for exploring the direction of SA-based research. It is necessary to validate if such a problem exists in the point cloud processing. Meanwhile, another concern of the the SA is its quadratic dependency on the input cardinality~\cite{wang2020linformer}, which limits its application to substantially down-sampled inputs~\cite{wang2018non}.

In this paper, we first perform qualitative and quantitative analyses to show that the SA tends to attend some fixed positions regardless of different queries. In other words, the attention is biased towards the learning of the query-independent information while suppressing the learning of query-dependent information. Based on this observation, we propose the unary-pairwise attention (\textbf{UPA}) by reformulating the SA to exploit query-independent and query-dependent information simultaneously while ensuring query-dependence. 
Specifically, given a query point and its nearest neighbors, the unary attention produces attention scores using absolute features, which ensures query independence of produced scores. The pairwise attention, by contrast, calculates attention scores using relative features to encode query dependence. Both attentions are permutation invariant; thus, it is fitting for 3D point cloud processing. The graphical description of UPA is shown in Fig.~\ref{fig: overview}. We show that the UPA outperforms the SA in various tasks through extensive experiments. Furthermore, the UPA brings systematical improvements for both standard and modern networks when it is integrated into networks as a compositional module.


The key contributions are summarized as follows: 
\begin{itemize}
\itemsep0em 
\item We conduct qualitative and quantitative analyses to show that the SA tends to attend fixed positions regardless of different queries.
\item We propose a new form of attention, the unary-pairwise attention (UPA), to enhance local feature learning of 3D point clouds.
\item We perform various experiments to demonstrate that the UPA consistently outperforms the SA across a range of tasks. Moreover, as a compositional module, the UPA systematically provides performance improvements for standard and modern networks.
\end{itemize}

\section{Related Work}
\hspace{\parindent}\textbf{Deep learning on 3D point clouds}. Owing to its unstructured nature, point clouds need to be projected onto regular grids to enable regular convolutions. Some methods convert point clouds into multi-view images~\cite{su2015multi, kanezaki2018rotationnet} while other methods voxelize 3D point clouds~\cite{maturana2015voxnet, zhou2018voxelnet,graham20183d, choy20194d}. The performance of image-based methods may heavily rely on the choice of projection planes whereas the memory costs of voxel-based methods grow cubically with the resolution. Besides, both types of methods lose fine-grained information due to projections. To overcome these issues, Qi~\etal{} propose PointNet~\cite{qi2017pointnet} that can directly operate on 3D point clouds. Subsequently, PointNet++~\cite{qi2017pointnet++} is proposed to tackle the local structure by applying PointNets to local subsets of point clouds.
Owing to its simplicity and effectiveness, PointNet++ becomes the key building block for recent studies~\cite{wang2019dynamic,li2018pointcnn,thomas2019kpconv,liu2019relation, xu2021paconv, xiang2021walk}. While their works focus on developing methods based on convolution, this study aims to develop a new operation based on the attention mechanism.

\textbf{Self-attention}. Transformers~\cite{vaswani2017attention} has revolutionized natural language processing and inspired vision researchers to apply the SA in image processing tasks~\cite{wang2018non, bello2019attention, parmar2018image,dosovitskiy2020image,carion2020end}. To further adapt the SA to specific applications, some works apply it locally~\cite{ramachandran2019stand,hu2019local}; some works make it more expressive~\cite{chen20182,yin2020disentangled}; some works improve efficiency~\cite{yue2018compact, cao2019gcnet}. The idea of the SA has also been introduced in point cloud/set processing~\cite{lee2019set}. PAT~\cite{yang2019modeling} develops a parameter-efficient variant whereas PointASNL~\cite{yan2020pointasnl} uses the SA to augment convolution-based networks. Some other works apply channel-wise modulation~\cite{wang2019graph,hu2020randla,zhao2020point} to exploit fine-grained details. By contrast, motivated by analyses on the SA, our approach aims to enhance the SA by explicitly modeling query-independent and query-dependent information simultaneously. Furthermore, unlike the SA that attends globally, our method operates locally to guarantee query dependence while being able to tackle the voluminous input.
\section{Method}
\label{sec: method}
\begin{figure}[t]
    \begin{center}
        \includegraphics[width=0.85\linewidth]{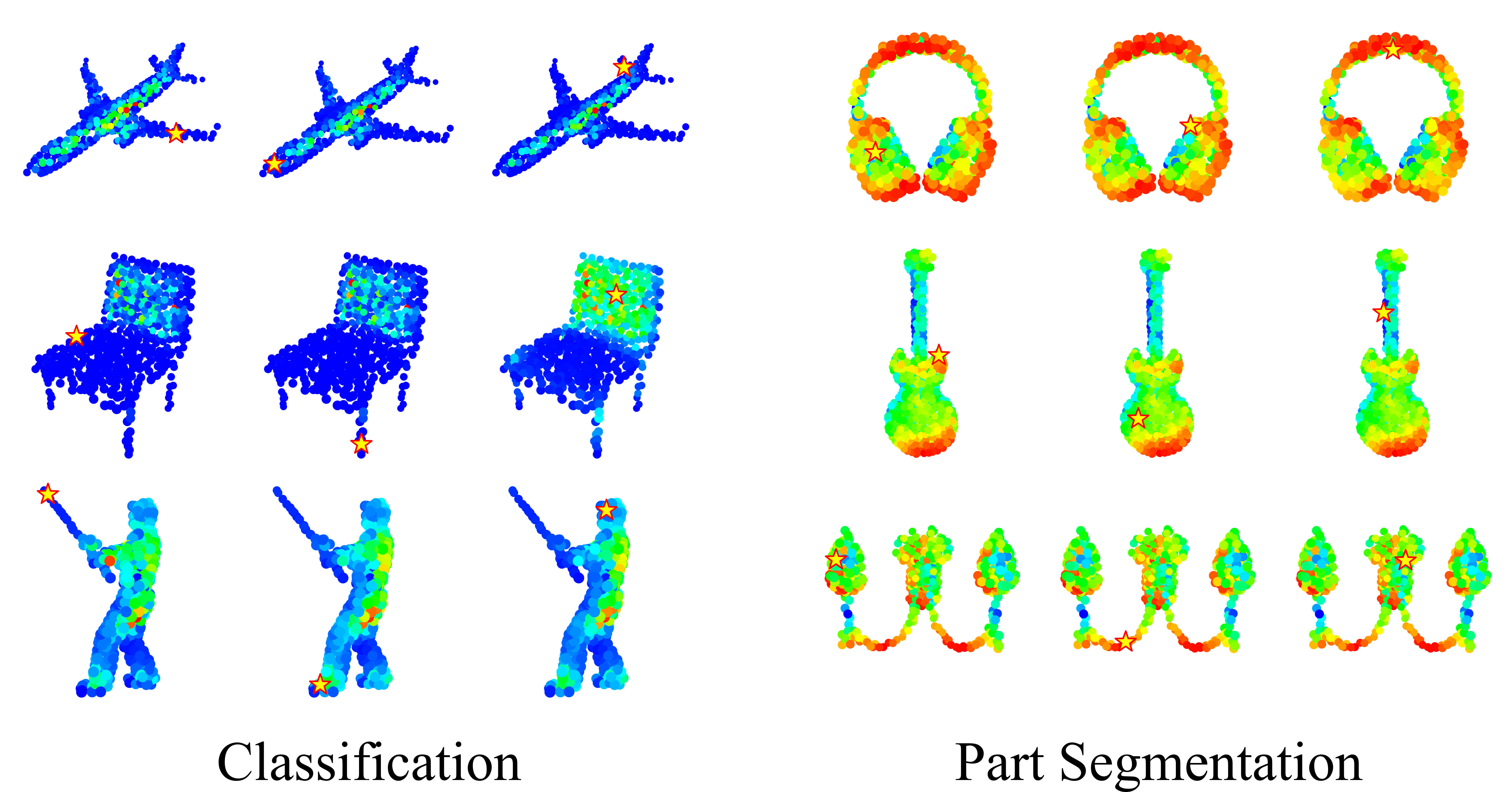}
    \end{center}
    \caption{\textbf{Attention maps generated by SA layers.} Analyses are performed on ModelNet (classification) and ShapeNet (part segmentation) datasets. Stars indicate query positions. Different query positions produce similar attention maps.}
    \label{fig: quali}
\end{figure}
\begin{table}[t]
    \centering
    \begin{tabular}{c|ccc|ccc}
         
         & \multicolumn{3}{c|}{ModelNet}
         & \multicolumn{3}{c}{ShapeNet}
         \\
         
         & OA
         & stage 1
         & stage 2
         & mIoU
         & stage 1
         & stage 2
         \\
         \hline
         PointNet++~\cite{qi2017pointnet++} + SA
         & 92.1
         & 0.094 
         & 0.047
         & 85.8
         & 0.000
         & 0.000
         \\
         PointNet++~\cite{qi2017pointnet++} + DNL~\cite{yin2020disentangled}
         & 92.1
         & 0.162
         & 0.147
         & 85.8
         & 0.000
         & 0.000
         \\
    \end{tabular}
    \caption{\textbf{Results of the quantitative analysis on SA and DNL layers.} Stage $n$ denotes the inserting position of SA and DNL layers after the $n$th set abstraction level~\cite{qi2017pointnet++}. Scores for each stage represent $\textrm{mJSD}=\frac{1}{N^{2}h}\sum_{i=1}^{N}\sum_{j=1}^{N}\mathrm{JSD}(\mathrm{AttMap_i}, \mathrm{AttMap_j})$, which measures the average similarity of attention maps over all query positions. $N$ denotes the number of input points, $h$ is the number of attention heads (8 for the classification and 1 for the segmentation), and $\mathrm{AttMap_i}$ and $\mathrm{AttMap_j}$ represent attention maps of points $i$ and $j$.}
    \label{tab:quanti}
\end{table}
This section begins by analyzing the SA and discussing the observed problems. Then, we present the formulation of the proposed UPA, and UPA blocks for seamless integration to existing networks as a compositional module.

\subsection{Analysis of Self-Attention}
Let $\mathbf{X}=\{\mathbf{x}_i\}_{i=1}^N$ denote the feature map of a point cloud where $N$ is the total number of points and $\mathbf{x}_i$ indicates a feature vector associated with a point. Then the SA can be defined as:
\begin{equation}
    \mathbf{y}_i = \sum_{j=1}^N 
        \mathrm{softmax}_j (\mathbf{q}_i^{\top}\mathbf{k}_j)\mathbf{v}_j
\end{equation}
where $i$ and $j$ index query and key elements, respectively. The query $\mathbf{q}_i=W_Q\mathbf{x}_i$, key $\mathbf{k}_j=W_K\mathbf{x}_j$, and value vectors $\mathbf{v}_j=W_V\mathbf{x}_j$ are linear transformations of the query point $\mathbf{x}_i$ and the key point $\mathbf{x}_j$. $W_Q$, $W_K$ and $W_V \in \mathbb{R}^{d_{in} \times d_{out}}$ are learned linear projections. $\mathbf{y}_i\in \mathbb{R}^{d_{out}}$ represents the output feature. $d_{in}$ and $d_{out}$ are input and output feature dimensions, respectively. $\mathrm{softmax}_j(\cdot)$ is applied to normalize dot product outputs between the query and the corresponding keys. As a result, the output $\mathbf{y}_i$ is a convex combination of value vectors.

While the SA is effective in point cloud recognition tasks, few studies investigate the behavior of attention maps generated by SA layers; thus, we provide qualitative and quantitative analyses of learned attention maps on classification (ModelNet40~\cite{wu20153d}) and part segmentation (ShapeNet~\cite{yi2016scalable}). To provide an intuitive understanding of the behavior of the SA, the qualitative analysis visualizes the attention maps of different query positions. Then, the quantitative analysis is performed to quantify the average similarity of all attention maps. We adopt point-averaged Jensen-Shannon Divergence (mJSD) as the similarity measure. To train SA layers, we adopt PointNet++~\cite{qi2017pointnet++}, which is the key building block for recent developments, as the backbone, and apply an SA layer after each set abstraction level~\cite{qi2017pointnet++}. The qualitative results are shown in Fig.~\ref{fig: quali}. Unexpectedly, attention maps are similar to each other although the query positions are different. Furthermore, as shown in Table~\ref{tab:quanti}, mJSD scores in SA layers are generally small, indicating that SA layers are prone to learning the global structure of point clouds by prioritizing query-independent information. In addition, we train DNL~\cite{yin2020disentangled} layers, in which the dot product is mathematically disentangled into query-independent and query-dependent terms, under the same setting as SA layers. Despite the improvements of mJSD scores in classification, DNL layers also degenerate into query-independent operators in a more challenging part segmentation task, revealing that the query-dependent information needs to be exploited more systematically. 

\subsection{Unary-Pairwise Attention}
\label{sec: upa}
Based on the observations from the above analysis, we propose the unary-pairwise attention (UPA) to handle the query-dependent and query-independent information simultaneously while ensuring the query dependence of attentions. We propose two distinct formulations that operate in parallel to optimize each component with minimal mutual interference. Furthermore, we apply the UPA to the local regions of query points because the SA tends to degenerate into a query-independent operator given a global receptive field.  In such a manner, attention outputs become query-dependent, which we find beneficial for point cloud processing. In addition, constraining operating scope also reduces the time/space complexity from quadratic to linear to the input cardinality, and thus enables UPA to be scalable to voluminous data. 

Formally, the general formulation of the UPA for a query $\mathbf{x}_i$ can be defined as:
\begin{equation}
    \mathbf{y}_i = 
        \sum_{\mathbf{x}_j \in \mathcal{N}(\mathbf{x}_i)}
            \mathrm{softmax} ( 
                f(\mathbf{x}_i, \mathbf{x}_j)
            )
            g(\mathbf{x}_j)
\end{equation}
where $\mathcal{N}(\mathbf{x}_i)$ is $k$-nearest neighbors of $\mathbf{x}_i$. $f$ is a relation function that measures the relationship between two inputs. $g:\mathbb{R}^{d_{in}}\mapsto\mathbb{R}^{d_{in}}$ is a transformation function that is implemented as a simple linear projection.

The exact form of the relation function $f$ is need-dependent. We introduce two distinct instantiations, one is responsible for exploiting the unary relation and the other for the pairwise one:
\begin{equation}
    f_u(\mathbf{x}_i, \mathbf{x}_j) = W_u\mathbf{x}_j
\end{equation}
\begin{equation}
    f_e(\mathbf{x}_i, \mathbf{x}_j) = W_e(\mathbf{x}_j - \mathbf{x}_i)
\end{equation}
The relation functions $f_u: \mathbb{R}^{d_{in}} \mapsto \mathbb{R}$ and $f_e: \mathbb{R}^{d_{in}} \mapsto \mathbb{R}$ map input features to scores for subsequent attention weight calculations.  
$W_u \in \mathbb{R}^{1 \times d_{in}}$ and $W_e \in \mathbb{R}^{1 \times d_{in}}$ are learned pointwise linear projections. Note that the transformation function $g$ is shared between two components to reduce the complexity.

For the unary relation, each neighbor $\mathbf{x}_j$ individually predicts a score for itself; thus, the generated attention map is independent of the pairwise relationship. On the other hand, the pairwise relation function $f_e$ maps the relative feature (relative to the query) to the score in which the pairwise interaction between the query and the neighbor is considered.

Like the SA, our formulation can be easily extended to the multi-head setting by arranging the relation function $f$ to predict $h$ scores and perform attention $h$ times in the corresponding subspace of input features. Then the output of each head is concatenated to compose the final outputs $\mathbf{y}_i = \mathrm{Concat}(\mathbf{y}_i^{head_1}, ..., \mathbf{y}_i^{head_h})$, where $\mathbf{y}_i^{head_h} \in \mathbb{R}^{d_{in} / h}$.


\begin{SCfigure}
  \includegraphics[width=0.45\textwidth]{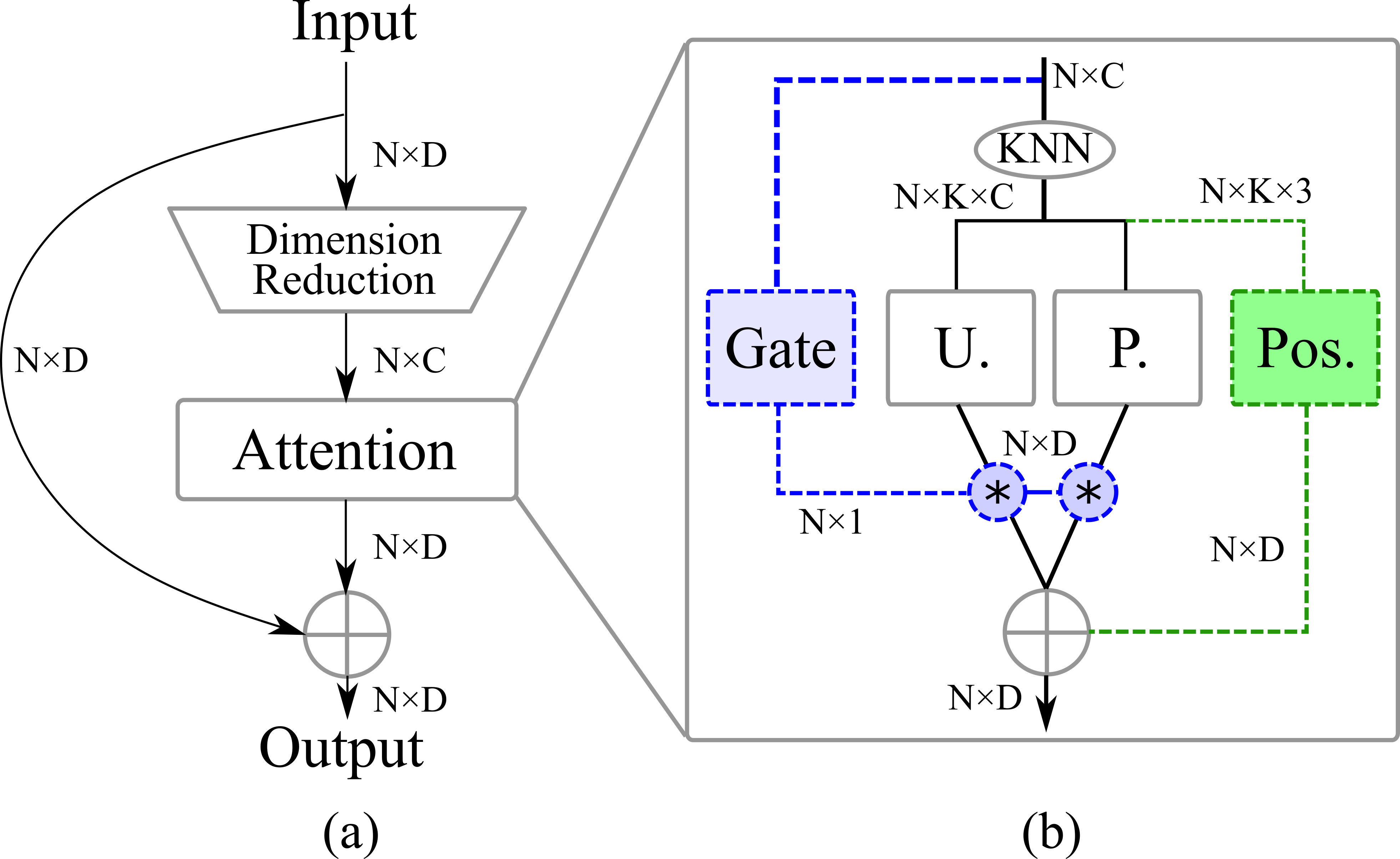}
  \caption{\textbf{Attention block designs.} \textbf{(a)} The general attention block used in the experiments. \textbf{(b)} The detail of the UPA block. The green branch is activated when the task is part segmentation whereas the blue one is activated when the task is classification or semantic segmentation.}
  \label{fig: block_design}
\end{SCfigure}

{\bf Design of the attention block.} The general block design is illustrated in Fig.~\ref{fig: block_design} (a). The block consists of a dimensionality reduction layer, which is an MLP, an attention layer, and a residual connection. The block receives input point clouds and produces the new feature vectors $\mathbf{z}_i \in \mathbb{R}^{d_{out}}$ that are the sum of input features and output of the attention layer. We construct the specific attention block by injecting the selected attention mechanism into the attention layer. The output of an attention mechanism is nonlinearly transformed and added to the input. For instance, in the case of the UPA, the output is calculated as: $\mathbf{z}_i = \mathbf{u}_i + \mathbf{e}_i + \mathbf{x}_i$, where $\mathbf{u}_i$ and $\mathbf{e}_i$ are nonlinearly transformed attention outputs $\alpha(\mathbf{y}_i^{unary})$ and $\beta(\mathbf{y}_i^{pairwise})$, respectively. $\alpha, \beta: \mathbb{R}^{d_{in}} \mapsto \mathbb{R}^{d_{out}}$ are component-wise MLPs. 

{\bf Task-specific UPA block designs.} The task-specific designs of the UPA block are illustrated in the Fig.~\ref{fig: block_design} (b). We additionally design the UPA block considering positional information~\cite{vaswani2017attention,carion2020end} for shape part segmentation tasks in which explicitly encoding 3D layout is found to be beneficial. Specifically, given 3D coordinates of a query point and its neighbor $\mathbf{p}_i, \mathbf{p}_j \in \mathbb{R}^{3}$, positional features are computed as $\mathbf{x}_{pos} = \delta (\mathbf{p}_j - \mathbf{p}_i)$, where $\delta: \mathbb{R}^3 \mapsto \mathbb{R}^{d_{in}}$ is an MLP consisting of two linear projections with a ReLU activation in between. Then, the positional encoding is produced by following the procedures of the unary attention treating $\mathbf{x}_{pos}$ as the input feature. The position encoding is shown as the green branch in Fig.~\ref{fig: block_design}. To tackle shape classification/scene segmentation, a gating mechanism is introduced to adaptively control the amount of information taken from each component. Specifically, each point predicts a score $s_i$ by linearly transforming the input feature $\mathbf{x}_i$ such that $\mathbf{z}_i = 
        \phi (s_i) \cdot \mathbf{u}_i + 
        \varphi (s_i) \cdot \mathbf{e}_i + 
        \mathbf{x}_i$.
We expect that an explicit gating is useful for enhancing/suppressing relevant/irrelevant information. For simplicity, we set $\phi$ and $\varphi$ as $\mathrm{sigmoid}(s_i)$ and $\mathrm{1-sigmoid}(s_i)$, respectively. The gating procedure is described graphically in Fig.~\ref{fig: block_design} (the blue branch).

\section{Experiments}
\label{sec: exp}
In this section, we present experimental results on the shape classification, part segmentation, and scene segmentation tasks. The performance of the UPA is compared with recent attention-based networks. Subsequently, we apply the UPA to various backbone networks to investigate its impact on standard and modern networks.

\textbf{Experimental settings}. We are particularly interested in comparing the relative performance improvements provided by various attentions. Specifically, experiments are performed by fixing the backbone architecture while altering attention layers. We choose the standard SA and DNL~\cite{yin2020disentangled} as baselines. Moreover, their localized variants (local-SA and local-DNL) are presented to quantify the direct impact of restricting receptive fields.
\begin{figure}[t]
     \begin{center}
        \includegraphics[width=0.85\linewidth]{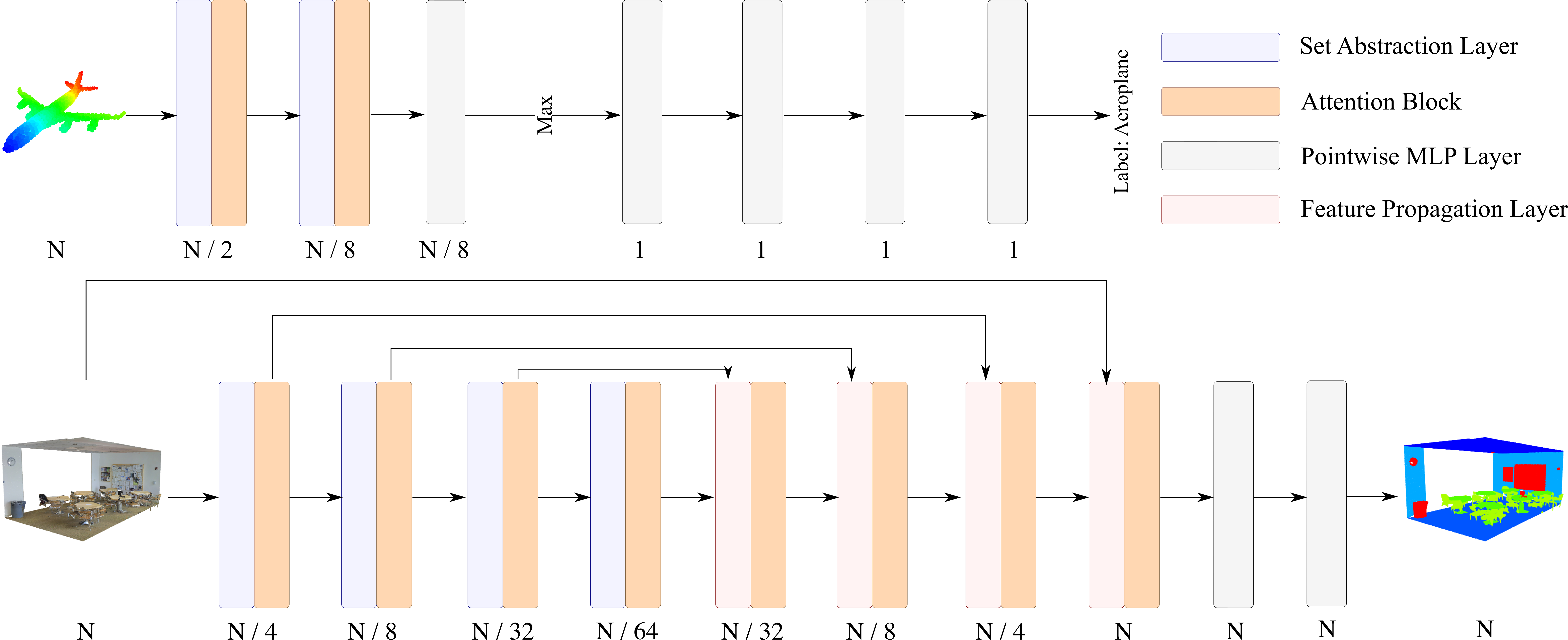}
    \end{center}
    \vspace{-9pt}
    \caption{\textbf{Task-specific architectures used in this study.} The upper model is used for the classification whereas the bottom one is used for the segmentation task. \textbf{N} represents the number of input points.}
    \vspace{-5pt}
    \label{fig: arch}
\end{figure}

We choose PointNet++~\cite{qi2017pointnet++} as the backbone because of its popularity as a building block in recent studies~\cite{li2018pointcnn,wu2019pointconv,liu2019relation,xu2021paconv}. The network architectures are illustrated in Fig.~\ref{fig: arch} and detailed configurations are reported in the supplementary material. Results using other backbones are presented in Sec.~\ref{subsec: integration}. To compare with recent attention-based methods, we also report the performance of Set Transformer~\cite{lee2019set}, PAT~\cite{yang2019modeling}, and PointASNL~\cite{yan2020pointasnl}. Other implementation details are reported in the supplementary material.

\vspace{-9pt}
\subsection{Shape Classification}
\hspace{\parindent}\textbf{Data}. We use ModelNet40~\cite{wu20153d} dataset, which contains 9,843 CAD models for training and 2,468 models for testing. We use the preprocessed point cloud data provided by~\cite{qi2017pointnet} for benchmarking. All inputs are normalized into a unit ball. We augment the input by random anisotropic scaling and random translation. Following~\cite{liu2019relation}, the input point number and features are set to 1,024 and 3D coordinates, respectively.  

\textbf{Results}. As shown in Table~\ref{tab:results}, UPA and local-DNL provide the greatest performance gain compared to others, revealing the effectiveness of modeling query-dependent and query-independent information simultaneously. Furthermore, UPA outperforms or is on par with powerful attention-based methods, showing its effectiveness as a compositional module. The usefulness of the restricted operating scope is verified as both local-SA and local-DNL outperform their global counterparts.
\begin{table}
    \begin{center}
        \begin{tabular}{c|ccc}
        Method & ModelNet40~\cite{wu20153d} & ShapeNet~\cite{yi2016scalable} & S3DIS~\cite{armeni20163d} 
        \\
        \hline
        Set Transformer~\cite{lee2019set}
        & 89.2
        &-
        &-
        \\
        PAT~\cite{yang2019modeling}
        & 91.7
        & -
        & 60.1
        \\
        PointASNL~\cite{yan2020pointasnl}
        & \textbf{92.9} 
        & 86.1
        & 62.6
        \\
        \hline
        PointNet++~\cite{qi2017pointnet++}
        & 90.7 
        & 85.1
        & 57.3
        \\
        SA (PointNet++)
        & 92.6 ($\uparrow 1.9$) 
        & 86.1 ($\uparrow 1.0$) 
        & 59.6 ($\uparrow 2.3$) 
        \\
        Local-SA (PointNet++)
        & 92.7 ($\uparrow 2.0$) 
        & \textbf{86.5} ($\uparrow \textbf{1.4}$) 
        & 60.1 ($\uparrow 2.8$) 
        \\
        DNL (PointNet++)
        & 92.1 ($\uparrow 1.4$)
        & 86.4 ($\uparrow 1.3$)
        & 57.9 ($\uparrow 0.6$)
        \\
        Local-DNL (PointNet++)
        & \textbf{92.9} ($\uparrow \textbf{2.2}$)
        & 86.3 ($\uparrow 1.2$)
        & 61.4 ($\uparrow 4.1$)
        \\
        \hline
        Ours (PointNet++)
        & \textbf{92.9} ($\uparrow \textbf{2.2}$) 
        & \textbf{86.5} ($\uparrow \textbf{1.4}$) 
        & \textbf{63.3} ($\uparrow \textbf{6.0}$) 
        \\
        \end{tabular}
    \end{center}
\caption{\textbf{The results of various point cloud understanding tasks.} The performance is measured using overall accuracy (OA), instance average IoU (mIoU), and class averaged IoU (mIoU) for classification, part segmentation, and scene segmentation, respectively.}
\label{tab:results}
\end{table}
\vspace{-9pt}

\subsection{Shape Part Segmentation}
\hspace{\parindent}\textbf{Data}. We use the ShapeNet Part dataset~\cite{yi2016scalable} to evaluate the performance on shape part segmentation. The dataset contains 16,880 models in which 14,006 are used for training and 2,874 for testing. Sixteen shape categories and 50 parts are included, each model being annotated with 2 to 6 parts. We use the data provided by~\cite{qi2017pointnet++} and take randomly sampled 2,048 points with the surface normal as input. The same augmentation strategy as the classification task is used. The voting~\cite{thomas2019kpconv, yan2020pointasnl,xu2021paconv} is applied as a post-processing step.  

\textbf{Results}. We use mean instance IoU as the performance metric~\cite{qi2017pointnet}. As reported in Table~\ref{tab:results}, UPA and local-SA achieves the best performance. Compared with the ones of the DNL and local-DNL, the UPA achieves the better performance, which verifies the suitability of the proposed formulations. Notably, except the SA, all attention variants outperform PointASNL, in which the standard SA is combined with convolutions, demonstrating their valid improvements over the SA.
\subsection{Scene Segmentation}
\hspace{\parindent}\textbf{Data}. We evaluate our models on Stanford large-scale 3D indoor spaces (S3DIS)~\cite{armeni20163d} dataset for scene segmentation. It contains six indoor environments including 272 rooms. Each point is annotated with one of 13 categories. We follow the data preparation procedure of PointNet~\cite{qi2017pointnet}. 
Specifically, each input point is represented by a 9-dim vector (XYZ, RGB, and normalized location as to the room). We train the model for about 50K iterations. We use Area five for testing and the others for training. 

\begin{figure}[t]
  \centering
  \includegraphics[width=0.8\linewidth]{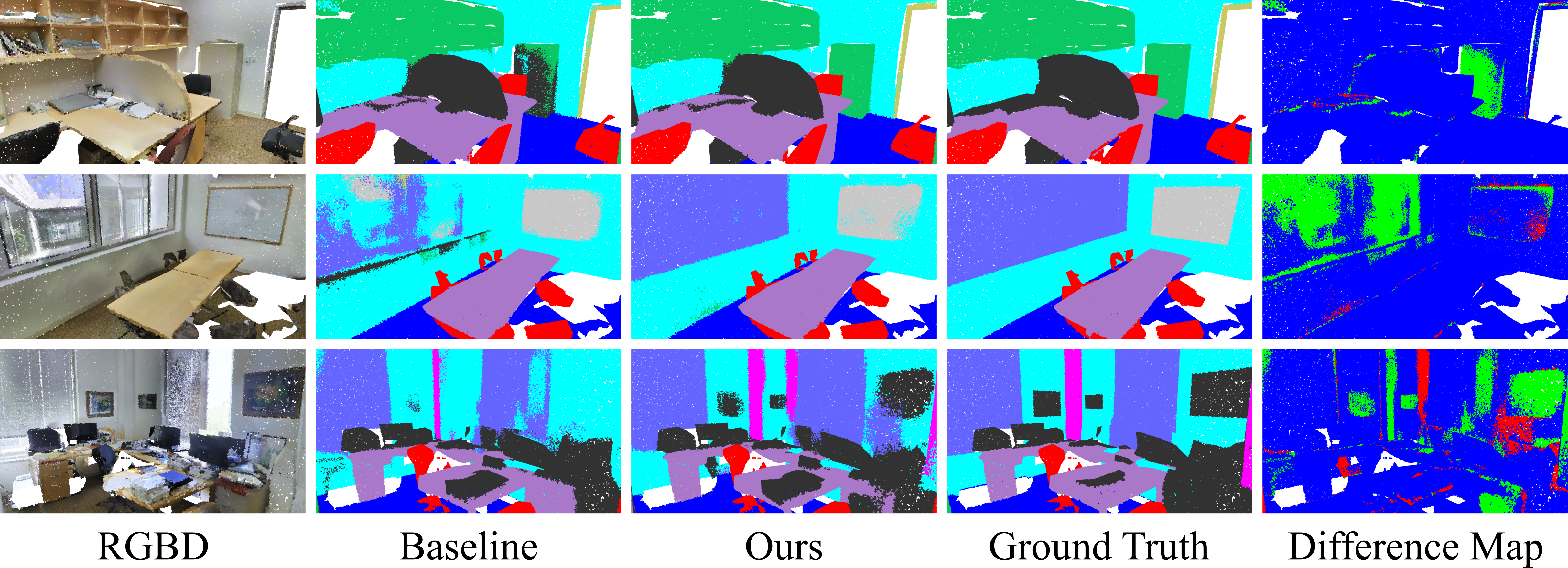}
  \caption{\textbf{Qualitative results of scene segmentation}. The last column shows the effect of the UPA: \textcolor{green}{green} indicates the points corrected by the UPA; \textcolor{red}{red} indicates wrongly modified points; \textcolor{blue}{blue} means unchanged points. The UPA generates smoother predictions compared to the baseline. For challenging scenes with a lot of clutter, the network equipped with UPA blocks is more boundary-aware and able to detect objects that the baseline fails to detect.}
  \label{fig: semseg}
\end{figure}

\textbf{Results}. As shown in Table~\ref{tab:results}, the UPA substantially improves the baseline by $6.0$ mIoU, outperforming other networks significantly in terms of the relative performance gain. The SA and local-SA successfully enhance the baseline; however, they have achieved lower relative gains. We conjecture that query-dependent information such as smoothness modeled by the pairwise term is especially crucial for scene understanding as many scenes are dominated by flat objects. As shown in Fig.~\ref{fig: semseg}, UPA obtains smoother predictions compared with the baseline. UPA successfully provides a greater improvement compared to local-DNL, which further verifies the usefulness of its pairwise term in scene understanding.

\vspace{-9pt}
\subsection{Integration with Modern Architectures}
\label{subsec: integration}
\begin{table}[t]
\centering
  \begin{tabular}{c|ccc|ccc}
        \multirow{2}{*}{Model} 
        &\multicolumn{3}{c|}{ShapeNet} 
        &\multicolumn{3}{c}{S3DIS}
        \\
        
        & Before
        & After
        &$\Delta$
        & Before
        & After
        &$\Delta$
        \\\hline
        PointNet~\cite{qi2017pointnet}
        & 83.7
        & 85.1
        & $\uparrow 1.4$
        & 41.1
        & 50.6
        & $\uparrow 9.5$
        \\
        PointConv~\cite{wu2019pointconv}
        & 85.7
        & 86.5
        & $\uparrow 0.8$
        & 62.8 
        & 64.9
        & $\uparrow 2.1$
        \\
        RSCNN~\cite{liu2019relation}
        & 86.2
        & 86.5
        & $\uparrow 0.3$
        & 62.3
        & 63.6
        & $\uparrow 1.3$
        \\
        PointASNL~\cite{yan2020pointasnl}
        & 86.1
        & 86.4
        & $\uparrow 0.3$
        & 62.6
        & 62.9
        & $\uparrow 0.3$
        \\
    \end{tabular}
    \caption{Results of adding UPA blocks to various backbones.}
    \label{tab:module}
\end{table}
We further investigate the effectiveness of UPA blocks by applying them to a wide range of existing networks. As shown in Table.~\ref{tab:module}, the UPA provides consistent improvements to all networks. In particular, the UPA successfully enhances PointASNL in which SA layers are used extensively, revealing that UPA is can provide additional benefits beyond the ones of the SA.

\vspace{-9pt}
\section{Design Analysis}
\begin{SCtable}
        \begin{tabular}{c|cc}
        Models & ShapeNet & S3DIS
        \\
        \hline
        Baseline
        & 85.1 & 57.3
        \\
        Unary 
        & 85.9 & 59.7
        \\
        Pairwise
        & 85.7 & 60.2
        \\
        Unary + Pairwise
        & 85.9 & 61.8
        \\
        Unary + Pairwise + Position
        & \textbf{86.1} & 60.8
        \\
        Unary + Pairwise + Gating
        & 85.7 & \textbf{63.3}
        \\
        \end{tabular}
\caption{\textbf{Results of the block component analysis.} Task-specific block designs effectively combine both attentions and improve the performance.}
\label{tab:component}
\end{SCtable}
We validate the design choices of the UPA in this section. Note that we do not perform voting in following experiments.
PointNet++ is used as the baseline throughout this section.
\subsection{Block Component Analysis}
As shown in Table~\ref{tab:component}, the unary attention is more effective in part segmentation while the pairwise one is more effective in scene segmentation. We suspect that query-independent feature is more useful in describing salient part boundaries whereas query-dependent features enforce smoothness in scenes in which flat objects dominate. The best performance is achieved after adding task-specific components to the block designs. However, the optimal design of the block is still an open question, which we leave to future work.


\subsection{Ablation Study}
\begin{SCfigure}
  \includegraphics[width=0.40\textwidth]{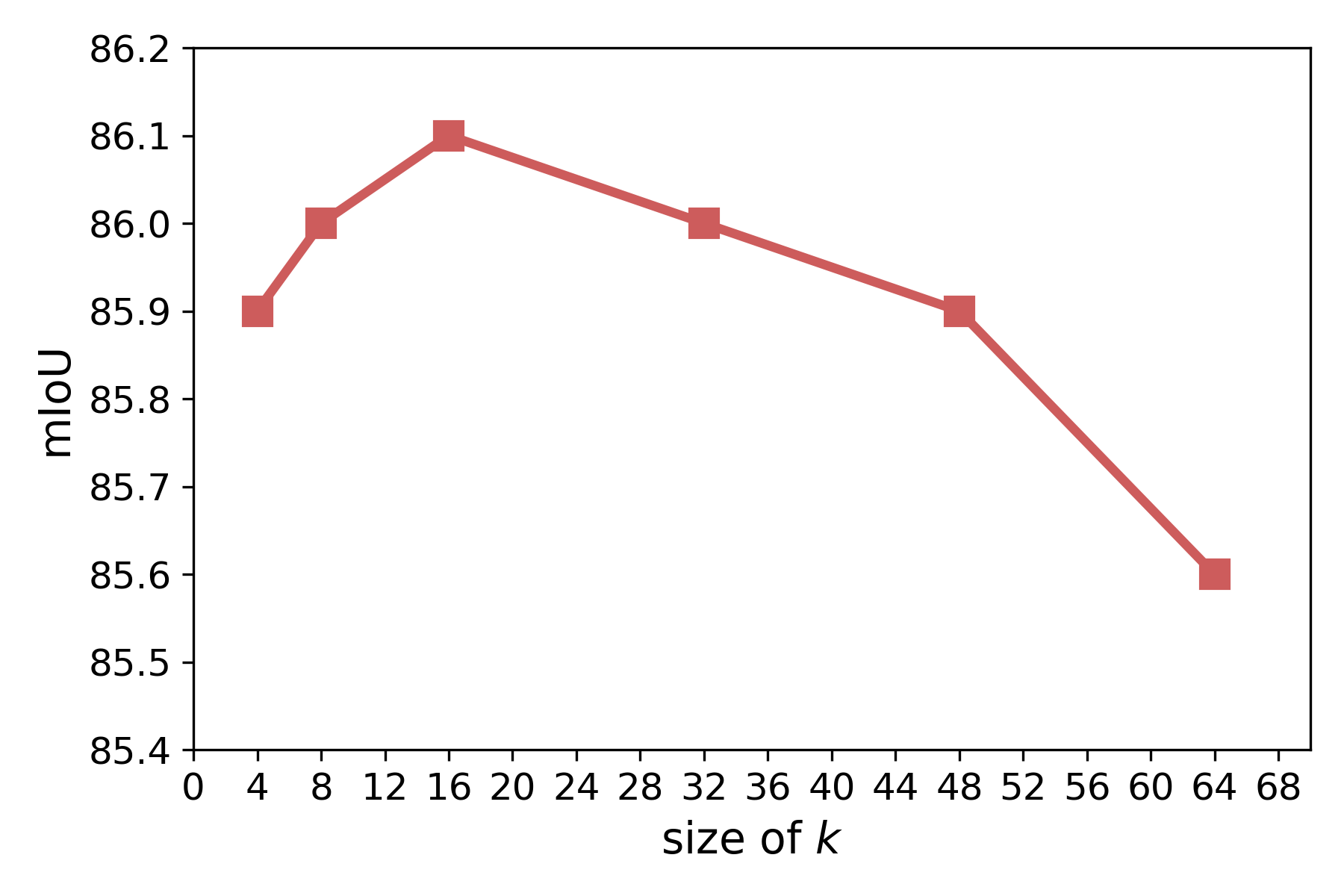}
  \caption{\textbf{Ablation study of neighborhood size $k$ on ShapeNet.} The performance reaches its peak at 16 and gradually starts to reduce when $k$ gets larger.}
  \label{fig: k_evolution}
\end{SCfigure}

\begin{table}[t]
    \centering
    \begin{tabular}{cc|cc|cc}
         
          \multicolumn{2}{c|}{\textbf{(a) Pooling  }}
         & \multicolumn{2}{c|}{\textbf{(b) Stage}}
         & \multicolumn{2}{c}{\textbf{(c) Compo. Arrangement}}
         \\\hline
         
          Baseline
         & 85.1
         & Baseline
         & 85.1
         & Baseline
         & 85.1
         \\

         Mean 
         & 85.7 
         & Stage 1
         & 85.5
         & Unary-Pairwise 
         & 86.0
         \\

         Max 
         & 85.3 
         & Stage 2 
         & 85.6
         & Pairwise-Unary 
         & 85.4
         \\

          Attention 
         & 86.1 
         & Stage 3
         & 85.8
         & Parallel 
         & 86.1
    \end{tabular}
    \caption{Results of the ablation study.}
    \label{tab:ablation}
\end{table}
\vspace{-4pt}

We choose part segmentation as the default task for ablation studies as we think that the task has sufficient complexity. We report the instance mIoU for each experiment. 

\textbf{Neighborhood size $k$}. As shown in Fig.~\ref{fig: k_evolution}, enlarging $k$ from 8 to 16 gradually improves the performance. However, mIoU starts to drop when $k$ gets larger. We conjecture that larger receptive fields contain information that is not helpful or even harmful for targeted tasks, thus complicating the optimization.

\textbf{Pooling method}. Average and Max improve performance, showing that fixed operations wrapped by our block are still beneficial. However, Attention offers more expressiveness as it achieves the best performance.

\textbf{Stage}. We examine the performance gain by adding a UPA block to each stage. Stage $n$ denotes the position of the block after $n$th set abstraction level. As shown in Table~\ref{tab:ablation}, the UPA influences the performance more when it is integrated into deeper stages.

\textbf{Component arrangement}. Here we compare the two arrangements of components: sequential and parallel. As shown in Table~\ref{tab:ablation}, the parallel arrangement outperforms all sequential ones, which verifies our design choice.

\vspace{-9pt}
\vspace{-9pt}
\section{Conclusions}
We propose the unary-pairwise attention (UPA) for enhancing 3D point cloud processing. Our analyses show that the standard self-attention (SA), which operates globally, is biased towards query-independent information, leaving query-dependent one not well exploited. As a result, the SA produces similar attention maps for different queries. Therefore, our new attention aims to jointly exploit both information while always being query-dependent by operating locally. Extensive experiments demonstrate that the UPA consistently outperforms the SA and other attentions, especially in the challenging task for which encoding query dependence appears useful. Moreover, equipped with the proposed UPA, the vanilla PointNet++ successfully outperforms or is on par with the state-of-the-art attention-based methods for various tasks. In addition, as a compositional module, the UPA successfully boosts the performance of various modern backbones, demonstrating its wide applicability.

\section*{Acknowledgement}
This paper is partially supported by a project, JPNP18010, commissioned by the New Energy and Industrial Technology Development Organization (NEDO) and JSPS Grant-in-Aid for Scientific Research (Grant Number 21K12042).

\bibliography{elsarticle-template}

\end{document}